%% file: arxiv3rd.tex
\pdfoutput=1
\documentclass{article}

\usepackage[final,nonatbib]{arxiv3rd}

\usepackage[utf8]{inputenc} 
\usepackage[T1]{fontenc}    
\usepackage{hyperref}       
\usepackage{url}            
\usepackage{booktabs}       
\usepackage{amsfonts}       
\usepackage{nicefrac}       
\usepackage{microtype}      

\usepackage{times}
\usepackage{epsfig}
\usepackage{graphicx}
\usepackage{amsmath}
\usepackage{amssymb}
\usepackage{tabulary}
\usepackage{algorithm}
\usepackage{subfigure}
\usepackage{algorithmic}

\title{Set-to-Set Hashing with Applications in Visual Recognition}

\begin{document}


\author{
   I-Hong Jhuo and Jun Wang\\
  Electrical Engineering, Columbia University\\
  \texttt{ihjhuo@gmail.com;jwang@ee.columbia.edu   } \\\\
}
\maketitle

\input{abstract}

\input{sec1}

\input{sec2}

\input{sec3}

\input{sec4}

\input{sec5}

\input{sec6}



{\scriptsize
{

}

\end{document}

%% file: abstract.tex
\begin{abstract}
Visual data, such as an image or a sequence of video frames, is often naturally represented as a point set. In this paper, we consider the fundamental problem of finding a nearest set from a collection of sets, to a query set. This problem has obvious applications in large-scale visual retrieval and recognition, and also in applied fields beyond computer vision. One challenge stands out in solving the problem---set representation and measure of similarity. Particularly, the query set and the sets in dataset collection can have varying cardinalities. The training collection is large enough such that linear scan is impractical. We propose a simple representation scheme that encodes both statistical and structural information of the sets. The derived representations are integrated in a kernel framework for flexible similarity measurement. For the query set process, we adopt a learning-to-hash pipeline that turns the kernel representations into hash bits based on simple learners, using multiple kernel learning. Experiments on two visual retrieval datasets show unambiguously that our set-to-set hashing framework outperforms prior methods that do not take the set-to-set search setting.

\end{abstract}

%% file: sec1.tex
\section{Introduction}
\vspace{-2mm}

Searching for similar data samples is a fundamental step in many large-scale applications. As the data size explodes, hashing techniques have emerged as a unique option for approximate nearest neighbor (ANN) search, as it can dramatically reduce both the computational time and the storage space. Successes are seen in areas including computer graphics, computer vision, and multimedia retrieval~\cite{Kulis_pami09,Sun_17,Wang_pami12,Wang_ariv15}. Hashing methods perform space partitioning to encode the original high-dimensional data points into binary codes. With the resulting binary hash codes, one can perform extremely rapid ANN search that entails only sublinear search complexity. 

Conventional hashing schemes concern point-to-point (P2P) search setting. They either depend on randomization and are data oblivious (represented by the classic Locality Sensitive Hashing -- LSH), or are based on advanced machine learning techniques to learn hashing functions that are better tailored to the specific data and/or label distribution. The latter includes unsupervised~\cite{Gong_cvpr11,Weiss_08}, semi-supervised~\cite{Wang_pami12,Wang_icml10}, and supervised hashing~\cite{Kulis_nips2009,Liu_cvpr12,Mu_cvpr10,Salakhutdinov_09,li16,Shen15,Zhang14}.

A natural generalization of the point-to-point search is set-to-set (S2S) search. For example, one can pose the facial image recognition problem as one that queries for a nearest subspaces to a given point~\cite{Wang_iccv13}. Indeed, there are several recent attempts, studying point-to-hyperplane search that is useful for active learning~\cite{Liu_icml12}, or subspace-to-subspace search~\cite{Basri_pami11} that models set-to-set search assuming linear structures in the sets. 

\begin{figure}
    \centering
    \subfigure[]
    {
    	\hspace{-6mm}
        \includegraphics[width=0.3\textwidth]{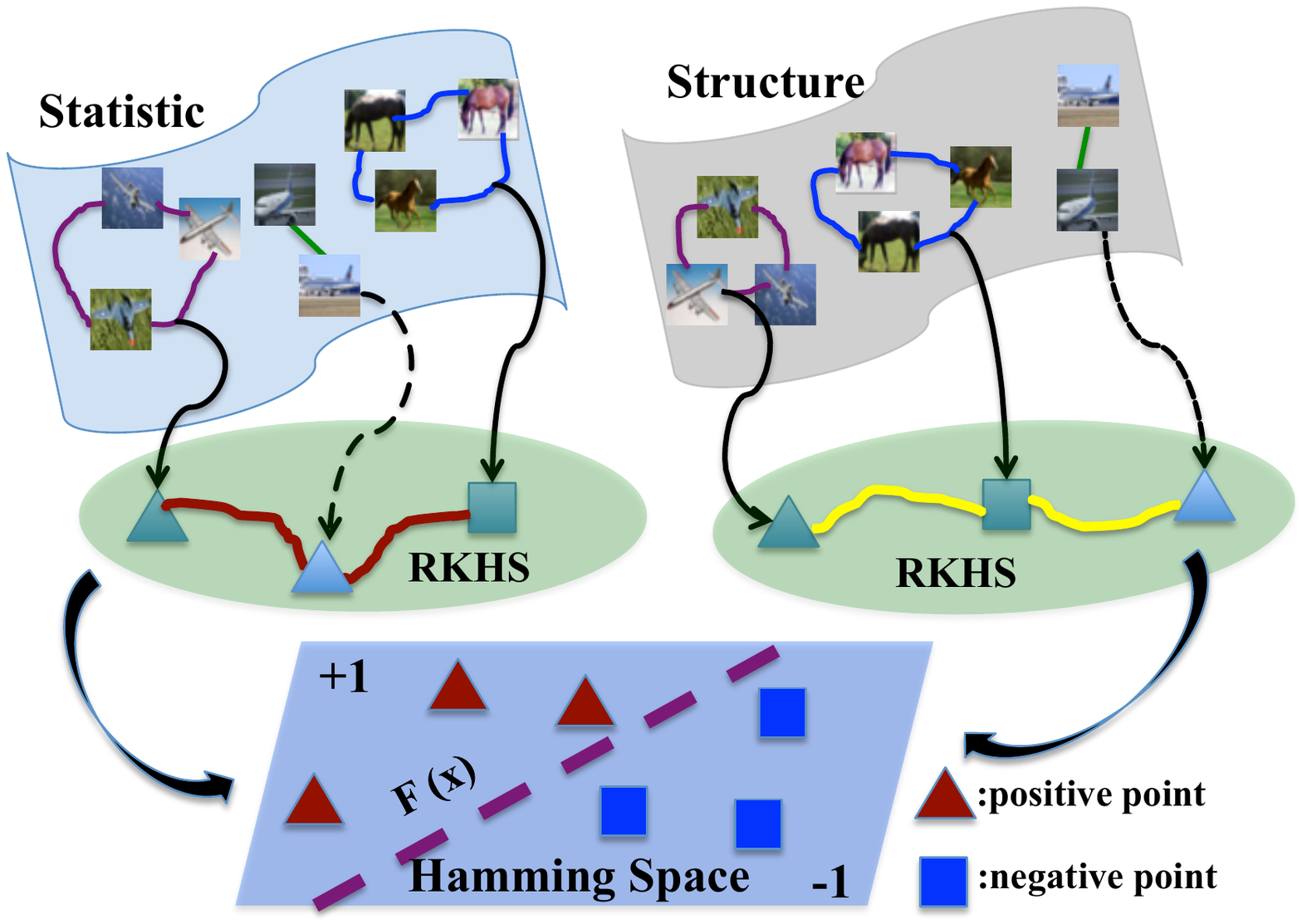}
    }
    \hspace{2mm}
    \subfigure[]
    {
        \includegraphics[width=0.45\textwidth]{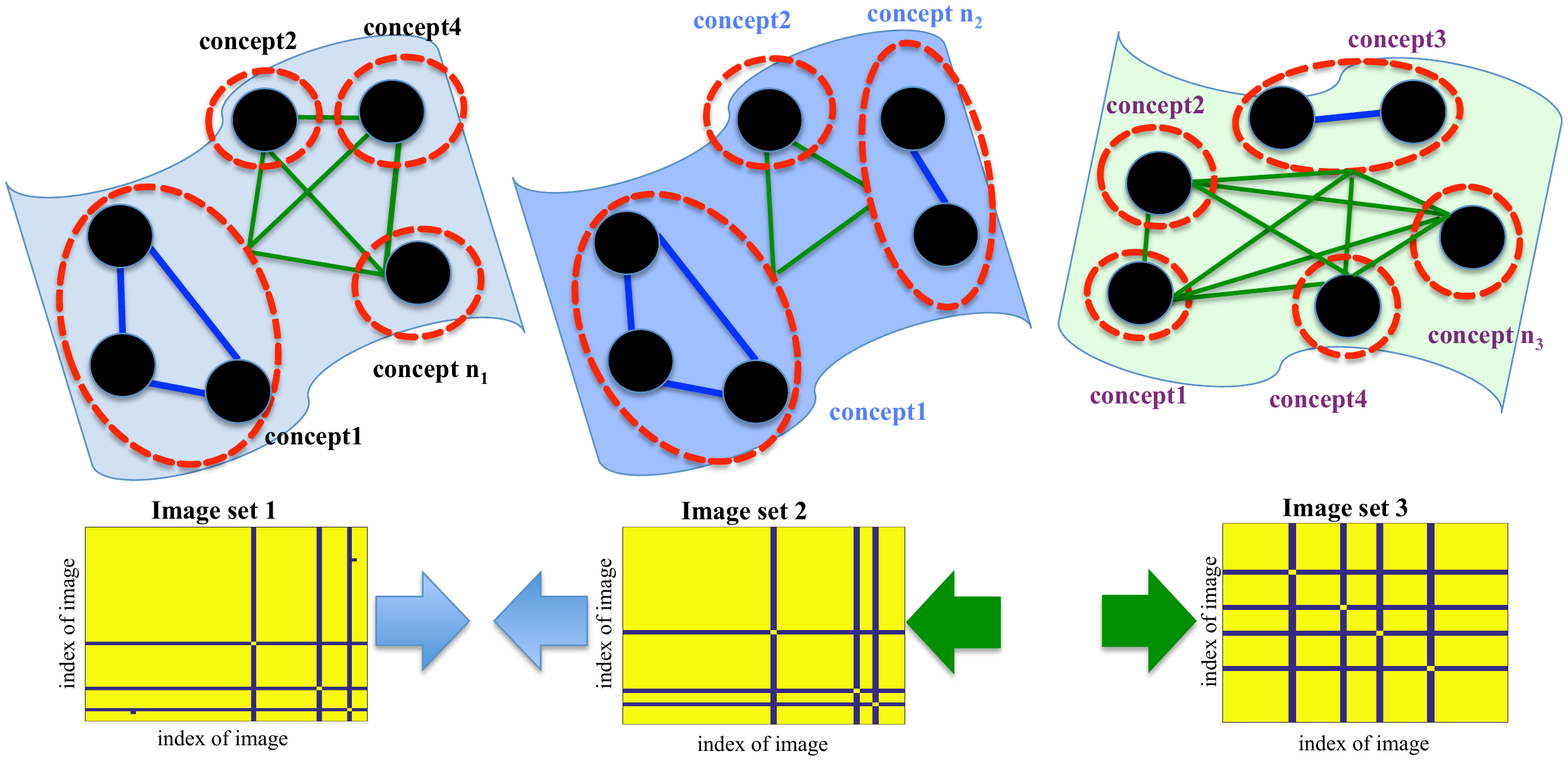}
    }
    \caption
    {
        (a) Illustration of the proposed ISH framework. First, statistical and structural information of the image sets are encoded. Second, appropriate kernel mappings are chosen to measure the similarities between image sets. A boosted algorithm based on the two kernels is used to construct the hash function. 
        (b) Extracting the structural information and measure the structural similarity. In the top row, graphs are constructed to represented individual sets: nodes are data points and graph weights indicate the similarities. Dense cliques (red circles) are then extracted to reveal the holistic structure in a set. Point sets with similar clique structures (sets 2 and 1 as shown) are assumed to have higher similarity. The bottom row shows the similarity matrices, in which yellow blocks indicate the high similarities among data points in each clique. Best viewed in color.
    }
    \label{fig:overview_structure}
\end{figure}

In this paper, we consider the set-to-set search problem in its full generality. This general setting finds applications ranging from video-based surveillance to 3D face retrieval from collections of 2D images~\cite{Berretti_pami10,Tuzel_cvpr07,Sivic_ivr05}.
Compared to specialized settings discussed above that come with natural notion of distance, a central challenge here is how to measure the distance/similarity between sets. We propose a similarity measure that captures both the statistical and structural aspects of the sets (Section 3). To learn the hash bits, we adopt \emph{dyadic hypercut} as a weak learner~\cite{Moghaddam_nips02} to derive a boosted algorithm that integrates both the structural and statistical similarities. The whole framework is illustrated in Fig.~\ref{fig:overview_structure}(a). 
In this paper, we focus on image applications, and hence coin the name Image Set Hashing (ISH). However, the core components of the proposed framework can be extended to generic scenarios.

%% file: sec2.tex
\section{Related Work}
\vspace{-3.5mm}
Here we discuss representative works in randomization-based hashing and learning-based hashing for P2P setting, and recent work on certain restricted S2S setting. Review of recent development of hashing techniques can be found in~\cite{Wang_ariv15,Wang_ariv14}.  

LSH hashing and variants are iconic randomization-based hashing schemes. They are simple in theory and efficient in practice, and flexible enough to handle various distance measures~\cite{Charikar_stoc02,Datar_scg04,Kulis_nips2009}. However, the data oblivious design of the LSH family methods often causes suboptimal recall-precision tradeoff curve. Hence, learning-based hashing schemes have been under active development recently. Generally, both the data and label information is fed into carefully designed learning pipeline to produce more adaptive and efficient hash codes. Depending on whether the label information is in use, these schemes are either unsupervised, e.g., spectral hashing~\cite{Weiss_08}, graph hashing, and ITQ (iterative quantization)~\cite{Gong_cvpr11}, or (semi-)supervised hashing, including~\cite{Lin_iccv13,Liu_cvpr12,Norouze_icml11,Mu_cvpr10,Wang_icml10}. Particularly, recent efforts have built more powerful hashing schemes on top of deep learning~\cite{Salakhutdinov_09,Salakhutdinov_07,Liong_cvpr15,Masci_pami13}. All these methods only deal with the P2P setting.



Study of the S2S setting started only very recently, and is mostly about image applications. Statistical or geometric assumptions are often made on the sets to facilitate representation. For example, statistical distribution of data points in each set can be assumed, and KL-divergence can be used to measure set similarity~\cite{Arandjelovic_cvpr05}. By comparison, point sets can also lie on linear subspaces or more general geometric objects~\cite{Cevikalp_cvpr10,Hu_cvpr11,Kim_pami07,Liu_cvpr14,Sun_14,Sun_15,Wang_cvpr12}. Among the representation schemes, representation based on covariance matrices has led to superior performances on image sets (video frames)~\cite{Wang_cvpr12,Lu_iccv13,Tuzel_cvpr07}. For instance, in~\cite{Li_cvpr15}, covariance matrix is used in such way, and similarity is then measured via kernel mapping and learning. Promising result has been reported, but the framework is restricted to cases when the query is a single point. For image applications specifically, the Set Compression Tree~\cite{Relja_14} compresses a set of image descriptors jointly (rather than individual descriptors) and achieve a very small memory footprint (as low as 5 bits). However, all existing representation methods do not account the holistic structural information; this may lead to incorrect similarity measurements on highly nonlinear data distributions. Our representation and hashing scheme is a first attempt to directly address the above problems in the general S2S setting.


%% file: sec3.tex
\section{Structural and Statistical Modeling}
\vspace{-2mm}
In this section, we detail how the structural and statistical information is extracted from the point sets, and how similarity between sets is measured. 

\subsection{Structure via graph modeling}
\vspace{-2mm}

The idea here is to use graph for discovering structures within data, and then measure the similarity via appropriate kernel mapping on graphs~\cite{Tenenbaum_00,Gartner03,Zhou_icml09}, Fig.~\ref{fig:overview_structure}(b) gives an example.  
To model data points within a set, we derive an affinity matrix $A$ based on quantized pairwise distances~\cite{Zhou_icml09}: if the distance is larger than a predefined threshold $\mu$, the corresponding affinity value in $A$ is set to $0$, and $1$ otherwise. We use $\mathbf x_i$'s to denote the point sets, and $A^i$'s to denote the corresponding affinity matrices thus constructed. With all the $A^i$'s at hand, the point set similarity is defined as: 
\begin{equation}
\scalebox{0.98}{$\
K_g(\mathbf x_i, \mathbf x_j)= \frac{\sum_{p=1}^{n_{i}}\sum_{q=1}^{n_{j}}A_{{ip}}A_{jq}g(x_{ip},x_{jq})}{\sum_{p=1}^{n_{i}}A_{ip}\sum_{q=1}^{n_{j}}A_{jq}}, 
\label{structure_fun}
$}
\end{equation}
where, $A_{ip}=1/\sum_{u=1}^{n_{i}}a_{pu}^{i}$, $A_{jq}=1/\sum_{v=1}^{n_{j}}a_{qv}^{j}$ and $g(x_{ip}, x_{jq})=\exp(- \gamma_g \parallel x_{ip}- x_{jq}\parallel^{2})$. $\gamma_g$ is a constant and $n_i$ and $n_j$ are the number of data points in $\mathbf x_i$ and $\mathbf x_j$, respectively.

To understand the captured structural information, each clique (formed by several $1$ elements) in $A^i$ can be regarded as one concept. If $A^i$ is an all-one matrix, all data points in one set belong to one concept and each data point set is considered as one data point. When $A^i$ is an identical matrix, each data point is independent, and no relation can be discovered. When $A^i$ is a clique-based matrix, data points can be clustered into cliques and $K_g$ is a clique-based graph kernel. In this way, we leverage the structural information for our set hashing.

\subsection{Statistical information}
\vspace{-2mm}
Covariance matrices have provided effective local region representation for visual recognition and human identification~\cite{Liu_cvpr14,Tuzel_cvpr07}. Intuitively, they describe the local image statistics. In this work, we use covariance matrices to depict the statistical variance of images within each set. 
Given $N$ image sets, $\mathcal X =\{(\mathbf x_1,l_1),\cdots, (\mathbf x_i,l_i), \cdots, (\mathbf x_N,l_N)\}$. $\mathbf x_i=\{x_{i1}, \cdots, x_{i,n_i}\}$ is an image set, where $x_{i,j} \in \mathbb{R}^{d}$ represents the $j$th d-dimensional feature vector in $\mathbf x_i$, and the set consists of $n_i$ images. $l_i$'s are the labels of each image set. Each image set is represented with a $d\times d$ covariance matrix:
\begin{equation}
\scalebox{0.98}{$\
C_i = \frac{1}{n_i}\sum_{j=1}^{n_i} (x_{ij} - \mathbf {\bar{x}}_i)(x_{ij} - \mathbf {\bar{x}}_i)^{\top},
\label{sementic_fun}
$}
\end{equation}
where $\mathbf {\bar x}_i$ is the mean feature vector within the set. The diagonal elements of $C_i$ represent the variance of each individual image feature, and the off-diagonal elements are their respective covariance. In this way, the covariance modeling can provide a desirable statistic for semantic variance among all images for an image set. Moreover, we use Gaussian-logrithm kernel~\cite{Jayasumana_cvpr2013} to map each covariance matrix of an image set into high dimension space, as follows: 
\begin{equation}
\scalebox{0.98}{$\
\begin{aligned}
	& K_s (\mathbf x_i, \mathbf x_j) = \phi(C_i)^{\top} \phi(C_j) \\
	& = \exp (-\left\|\log(C_i)-\log(C_j)\right\|_F^2 /2\gamma_{s}^{2}) ,\\
\end{aligned}
$}
\label{logkernel_fun}
\end{equation}
where $\gamma_s$ is a positive constant, which be set to the mean distances of training points. Kernel matrix $K_s$, is for the image sets in Riemannian space and $\parallel\cdot\parallel_F$ denotes the matrix Frobenius norm. In the end, each pair of image sets $\mathbf x_i$ and $\mathbf x_j$ are mapped by the $K_g$ and $K_s$ kernel functions into a high dimensional space.

%% file: sec4.tex
\section{Image Set Hashing}
\vspace{-2mm}
\subsection{Learning framework}
\vspace{-1mm}
Suppose we have an image set dataset $\mathcal{X} = \{\mathbf x_{i},l_i\}$. The goal of hashing is to generate an array of appropriate hash functions $h: \mathbb{R}^d \mapsto \{0,1\}$ by a designed function $\Psi$ and each bit is constructed by $h(\mathbf x) =\mathrm{sign} (\Psi(\mathbf x))$. However, there is no straightforward way to generate hash codes for each image set. Inspired by~\cite{Vemulapalli_cvpr13,Li_cvpr15,Wang_cvpr12} , we develop a mapping framework to construct hash codes for image sets in a common Hamming space based on multiple kernels. Kernel methods~\cite{Hamm_08,Jayasumana_cvpr2013,Vemulapalli_cvpr13,Wang_cvpr12} are known to capture and unfold rich information in data distribution. After the mapping process, we can generate hash codes in a Hamming space by simultaneously considering the structural and statistical information and iteratively maximize the discriminant margins based on multiple kernels learning.  

\subsection{Weak learners with boosting algorithm for hash functions}
\vspace{-1mm}
Since a multi-class classification problem can always be treated as an array of two-class problems by adopting one-against-one or one-against-all strategies, we design a boosting algorithm to learn binary splits for constructing hash functions. Specifically, we consider \textit{dyadic hypercut}~\cite{Moghaddam_nips02} with multiple kernel functions. A dyadic hypercut $f$ is generated by a kernel with a pair of different labels in training samples. Specifically, $f$ is parameterized by positive sample $\mathbf x_{a}$ , negative sample $\mathbf x_b$ , and kernel functions $\{K_m\}$,i.e., $m$ indicates statistical kernel ($K_s$) or structural kernel ($K_g$), and can be represented as follows:
\begin{equation}
\scalebox{0.98}{$\
f(\mathbf x)=\mathrm{sign}(K_m(\mathbf x_a,\mathbf x)-K_m(\mathbf x_b,\mathbf x)+\varepsilon), 
\label{dyadic_fun}
$}
\end{equation}
where $\varepsilon \in \mathbb{R}$ is a threshold. The size of the totally generated weak learner pool is $|{f}|= M \times n_a \times n_b$, where $M$, $n_a$ and $n_b$ are the numbers of kernels, positive training samples and negative training samples, respectively. With an efficient boosting process, we iteratively select a subset of weak learners by considering the learning loss. 

Note that the learning process may be susceptible to overfitting when $|f|$ is large.  To alleviate this issue, we adopt a boosting algorithm~\cite{Freund_CLT95,Moghaddam_nips02} to combine a number of weak splits (weak learners) into a strong one. Specifically, we iteratively select the discriminant weak learners generated from multiple kernels via maintaining a weighted distribution $w^{t}$ over data. Each iteration $t$ produces a weak hypothesis $f(\mathbf x): \mathbf x\rightarrow\{+1, -1\}$ and a weighted error $\delta^t$. The learning algorithm is aimed at selecting weak learner $f^{t}$ for minimizing $\delta^t$ followed by updating next distribution $w^{t+1}$.  We adopt \textit{exponential loss}~\cite{Freund_CLT95} and minimize the loss function to select the best weak learner $f^t$ at iteration $t$ and the best weak learner is computed as: 
\begin{equation}
\scalebox{0.98}{$\
f^t = \min_{f} \sum_{i=1}^{N} w_i^t \mathrm{exp} (-l_i f(\mathbf x_i)), 	
$}
\end{equation}
where $w_i^t$ indicates the weight of $\mathbf x_i$ at iteration $t$. Once obtaining the best weak learner, we update the data distribution based on weighted errors. The linear combination of weak learners, i.e., a strong split, is computed as follows: 
 \begin{equation}
 \scalebox{0.98}{$\
F(\mathbf x)=\mathrm{sign}(\sum_{t=1}^T\lambda^{t}f^{t}(\mathbf x)),
\label{strong_fun}
$}
 \end{equation}
where $\lambda^t = \frac{1}{2}\log \frac{1-\delta^t}{\delta^t}$. At each iteration $t$, $F = \sum_{\tau=1}^{t-1} \lambda^{\tau}f^{\tau} $ is a linear combination of the $(t-1)$ weak learners.

\subsection{Objective Function}
\vspace{-1mm}
With the designed hash functions, the following are desired properties of the hash codes: (1) Each hash value is independent of the binary representation for each sample. (2) When samples are close to each other in feature space (e.g. with similar distributions), the hash codes should induce similar hash values with a small Hamming distance. (3) In the resulting Hamming space, different contents of samples should have different hash codes, which push different samples of categories as far as possible, meanwhile gather the samples of the same category close to each other. Based on the criteria, we derive our multiple kernel hashing as follows:

\begin{equation}
\scalebox{0.98}{$\
\begin{aligned}
	& \min_{H_{q},H_{r}}  \alpha D_s + \beta D_c + \nu_1 \sum_{\substack{r' \in \{1:R\}\\
                  i \in \{1:N\}}} 
                  \Omega \\
  s.t. &  \quad H_q^{r'i}=\mathrm{sign}(\sum_{t=1}^{T} \lambda_{t} f_q^{tr'}(\mathbf x_i)), \forall  i \in \{1:N\}, \forall r' \in \{1:R\} \\
       &  \quad H_r^{r'i}=\mathrm{sign}(\sum_{t=1}^{T} \lambda_{t} f_r^{tr'}(\mathbf x_i)), \forall  i \in \{1:N\}, \forall r' \in \{1:R\}, \\
\end{aligned}
  \label{obj_fun}
  $}
\end{equation}
where $H_{*}^{r'i}$ is the hash value of the $i$th image set using the $r'$th strong split, i.e., hash function, and $r$ and $q$ represent the retrieval and query sets in the training process. $R$ is the number of the splits and $\Omega=\sum_{t=1}^T (\lambda_q^{t} f_{q}^{tr'} (\mathbf x_i) + \nu_2 \lambda_r^{t} f_{r}^{tr'} (\mathbf x_i))$. $f_{*}^{r'}$is the $r'$th strong split generated from the number of $T$ weighted weak learners (in eq. (\ref{strong_fun})), $\lambda_{*}^{t}$ is the coefficient of weak learners trained via a boosting algorithm, $\nu_1$ and $\nu_2$ are constant parameters. 

The minimization of the first two terms, i.e., $D_s$ and $D_c$, tend to find an optimal difference of the distance between intra- and inter- categories, which capture the discriminative property among all the samples and are defined in eq. (\ref{ds_fun}) and eq. (\ref{dc_fun}), and $d(\cdot, \cdot)$ indicates distance measure in Hamming space. In addition, $D_c$ can refine the hash codes generated from the training image sets (q and r parts) by maximizing the separability of category algorithm in~\cite{Rastegari_eccv12}. Simultaneous consideration of the two distance functions helps to minimize the within-category distances and meanwhile maximize the between-category dis. By formulating the structural and statistical information with multiple kernels for our objective function, the image set hashing becomes more robust and discriminative.


\begin{equation}
\scalebox{0.98}{$\
	D_s = \sum_{(m, n) \in \mathcal M} d(H_*^{m}, H_*^{n}) - \nu_3 \sum_{(m, n) \in \mathcal C} d(H_*^{m}, H_*^{n})
	\label{ds_fun}
$}
\end{equation}

\begin{equation}
\scalebox{0.98}{$\
	D_c = \sum_{(m, n) \in \mathcal M} d(H_q^{m}, H_r^{n}) - \nu_4 \sum_{(m, n) \in \mathcal C} d(H_q^{m}, H_r^{n}),
	\label{dc_fun}
	$}
\end{equation}
where $\mathcal M$ and $\mathcal C$ are represented as intra- and inter-category, $H_*$ can be $H_q$ or $H_r$, and $\nu_3$ and $\nu_4$ are the pre-computable constant parameters to balance the intra-catetory and inter-category scales. 


\algsetup{indent=2em}
\renewcommand{\algorithmicrequire}{\textbf{Input:}}
\renewcommand{\algorithmicensure}{\textbf{Initialize:}}
\newcommand{\alm}{\ensuremath{\mbox{\sc Image Set Hashing}}}
\begin{algorithm}[t]
\algsetup{linenosize=\tiny}
\scriptsize
\caption{Image Set Hashing }\label{alg:alm}
\begin{algorithmic}[1]
\REQUIRE a set of training image sets, $\mathcal{X} = \{\mathbf x_{i},l_i\}$ is divided into q and r two training image set parts, where $\mathbf x_i = \{x_{ij}\}_{j=1}^{n_i} \in \mathbb R^d$, $i \in \{1,2,\cdots, N\}$, $l_i = \{1,\cdots,L\}$.\\ 

\ENSURE Compute kernel matrices for q training image sets, i.e., $(K_{m})_q$, by using the kernel functions ($K_g$ and $K_s$) according to eq. (\ref{structure_fun}) and eq. (\ref{logkernel_fun}); Similar to r training image sets, $(K_{m})_r$\\

\STATE $V_q \in \mathbb R^{N \times R}$, $V_r \in \mathbb R^{N \times R}$ $\leftarrow$ kernel PCA with statistical kernels for $K_q$ and $K_r$, respectively
\STATE $H_{q} \leftarrow \mathrm{sign}(V_q^{\top}(K_s)_q)$
\STATE $H_{r} \leftarrow \mathrm{sign}(V_r^{\top}(K_s)_r)$

\textbf{Optimization:}

\WHILE{not converged}
\STATE Optimize $H_q$, $H_r$ with eq.(\ref{ds_fun})
\STATE Train $R$ splits by weak learner selection in e.q.(\ref{strong_fun}) on q kernels $(K_{m})_q$ by using $H_r$ as training labels, and inversely train another $R$ splits on r kernels $(K_{m})_r$ by using $H_q$ as training labels
\STATE $H_{q} \leftarrow \mathrm{sign}(\sum_{r'=1}^R F_{q}^{r'}(\mathbf x))\leftarrow (K_{m})_q$
\STATE $H_{r} \leftarrow \mathrm{sign}(\sum_{r'=1}^R F_{r}^{r'}(\mathbf x))\leftarrow (K_{m})_r$
\STATE Optimize $H=[H_q,H_r] \in \lbrace 0,1\rbrace^{R \times 2N}$ with eq.(\ref{dc_fun}) \\

\STATE  Train $R$ splits by weak learner selection in e.q.(\ref{strong_fun}) on q kernels $(K_{m})_q$ by using $H_r$ as training labels, and inversely train another $R$ splits on r kernels $(K_{m})_r$ by using $H_q$ as training labels \\
\STATE  Check the convergence condition
\ENDWHILE \\
\STATE  \textbf{Output:} $\sum_{r'=1}^R F_{q}^{r'}(\mathbf x)$ and $\sum_{r'=1}^R F_{r}^{r'}(\mathbf x)$ for query and database encoding in the testing process, respectively.

\end{algorithmic}
\end{algorithm}

\subsection{Optimization}
The objective function optimization problem~\ref{obj_fun} is a typical nonsmooth, nonconvex multiple variable minimization problem. We derive an iterative block coordinate descent algorithm~\cite{Tseng_01} for the optimization. Algorithm~\ref{alg:alm} gives the entire algorithm. Here, we highlight several critical steps in our algorithm. We first compute the kernel matrices $K_g$ and $K_s$ with eq. (\ref{structure_fun}) and eq. (\ref{logkernel_fun}) for q training and r training image sets. After the kernel computation, we adopt kernel PCA~\cite{Scholkopf_97} for the q and r two parts to obtain the initial hash codes, i.e., $H_q$ and $H_r$, based on their statistical kernels in Step 1 to Step 3. Next, we update the $H_q$ and $H_r$ codes by optimizing eq. (\ref{ds_fun}) for seeking the discriminability and utilize an efficient subgradient descent algorithm~\cite{Rastegari_eccv12} for the binary optimization (the optimization algorithm gives the generated code with two properties: sample-wise balance and bit-wise balance.). In Step 6, we use the updated hash codes, $H_q$ and $H_r$, to train $R$ two-class strong splits based on multiple kernels. More specifically, we adopt cross-training strategy~\cite{Rastegari_icml} by using the hash codes, $H_r$, as training labels to train the strong splits with q training image sets and similar process to $H_q$ hash code with r training image sets. After that, we update the current hash codes, $H_q$ and $H_r$ by using the learned strong splits based on multiple kernel learning. In order to improve the discriminability, we combine $H_q$ and $H_r$ together to refine the learned hash codes with eq. (\ref{dc_fun}). The process is then repeated. Convergence typically occurs within few outer iterations. Once we obtained the two strong split models ($\sum_{r'=1}^R F_{q}^{r'}(\mathbf x)$ and $\sum_{r'=1}^R F_{r}^{r'}(\mathbf x)$ in Algorithm 1), we can adopt them to generate query and database  hash codes in the testing process, respectively.

{\em Complexity Analysis:} The time complexity for training ISH is bounded by $\mathcal{O}(RS^2+(d+f)RS^3)$, where $S$ is the number of image sets, $f$ is number of dyadic-cuts, and $d$ indicates the dimension of samples and $N$ $\gg$ $S$, i.e., $N$ is the number of images. The hash look-up time of is $\mathcal{O}(RT)$, where $R$ is the number of bits and $T$ is the number of iterations in boosting algorithm. The major computational cost lies in computing kernels; however, it can be speeded up greatly using kernel approximation techniques or pre-computed kernel representation. For the practical encoding time, we only need $7.02 \times 10^{-5}$ seconds to encode one bit, which is faster than the HER method~\cite{Li_cvpr15} with $1.37 \times 10^{-4}$ seconds.

\begin{table*}[t]
\vspace{2mm}
\footnotesize
\centering
    \caption{The evaluation results measured by Mean Average Precision on the CIFAR-10 (60K) dataset. The numbers of training samples for P2P and S2S hashing are set to be $1000$ points and $195$ image sets, respectively. Three methods, LSH, SH, and SSH, are trained with images, while the KLSH, KSH, HER and ISH methods are trained with image sets.}
    \label{tb:cifar}
    \vspace{-1mm}
    \setlength{\tabcolsep}{8pt}
    \begin{center} 
        \begin{tabular}{|p{1cm}|p{1.1cm}|p{1.1cm}|p{1.1cm}|p{1.1cm}|p{1.1cm}||p{1.1cm}|p{1.1cm}|p{1.3cm}|}
            \hline
            &\multicolumn{6}{c|} {Compared Methods} & Our Method \\ \hline\hline
            bits  &  LSH~\cite{Indyk_98}      &   SH~\cite{Weiss_08}  & SSH~\cite{Wang_pami12} &  KLSH~\cite{Kulis_nips2009} & KSH~\cite{Liu_cvpr12}             & HER~\cite{Li_cvpr15}            &  ISH \\ \hline \hline
            8bits     &  $0.1063$          & $0.1279$            & $0.1165$       &  $0.1067$  &    $0.2363$ & $0.2284$   & $\textbf{0.2822}$\\
            12bits    &  $0.1073$          & $0.1330$            & $0.1416$       &  $0.1210$  &    $0.2486$ & $0.2672$   & $\textbf{0.3069}$\\
            24bits    &  $0.1086$          & $0.1317$            & $0.1512$       &  $0.1420$  &    $0.2680$ & $0.2772$   & $\textbf{0.3159}$\\
            32bits    &  $0.1194$          & $0.1322$            & $0.1574$       &  $0.1501$  &    $0.2818$ & $0.2937$   & $\textbf{0.3292}$\\
            48bits    &  $0.1105$          & $0.1352$            & $0.1629$       &  $0.1622$  &    $0.3003$ & $0.3154$   & $\textbf{0.3334}$\\
            
            \hline
        \end{tabular}
    \end{center}
\end{table*}

\begin{figure*}
  \hspace{-3mm}
  \begin{minipage}{0.3\textwidth}
    {
    \epsfig{file=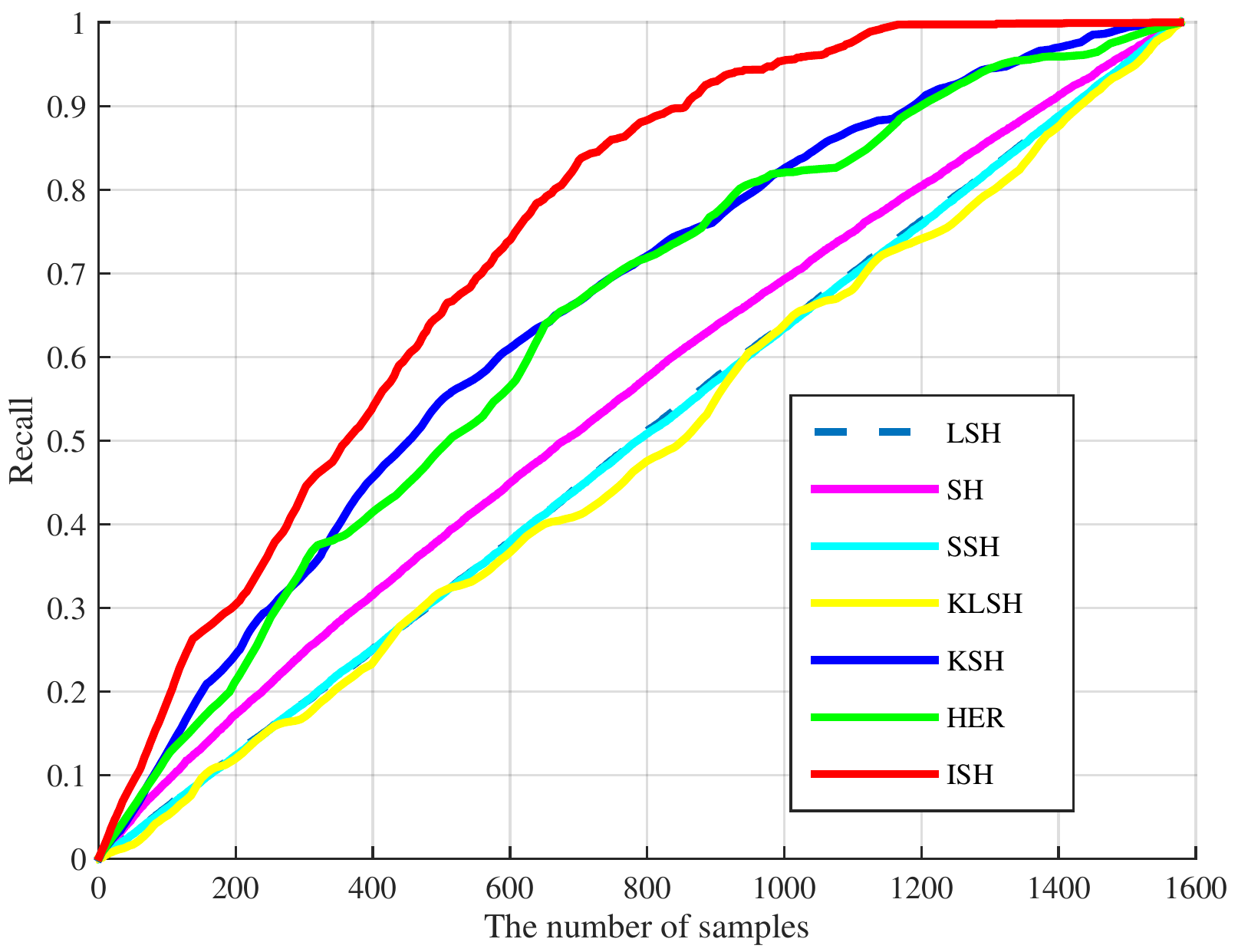, height=1.5in, width=2.1in }
    }
    \vspace{-7mm}
\caption{The evaluation results by mean recall curves for Hamming ranking using $24$ bits on the CIFAR-10 dataset. The number of retrieved samples is up to $1600$. The figure is best viewed in color.}
\label{fig:re_cifar}
  \end{minipage}
    \hspace{3mm}
  \begin{minipage}{.3\textwidth}
		{
	\epsfig{file=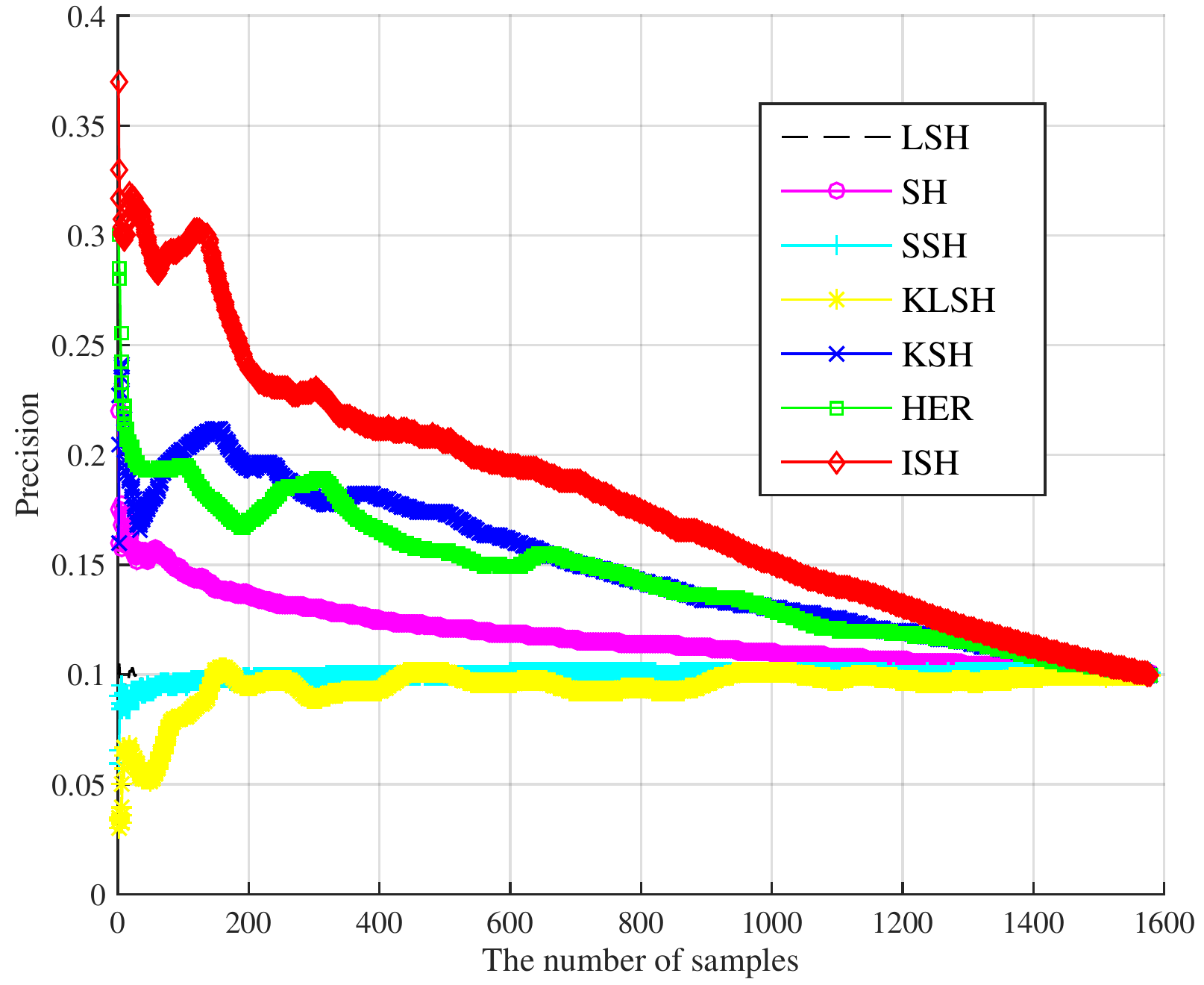, height=1.5in, width=1.9in }
		}
		\vspace{-7mm}
\caption{The evaluation results by mean precision curves for Hamming ranking using $24$ bits on the CIFAR-10 dataset. The number of retrieved samples is up to $1600$. The figure is best viewed in color.}
\label{fig:pr_cifar}
  \end{minipage}
  \hspace{4mm}
  \begin{minipage}{.3\textwidth}
    {
    \vspace{-2mm}
	\epsfig{file=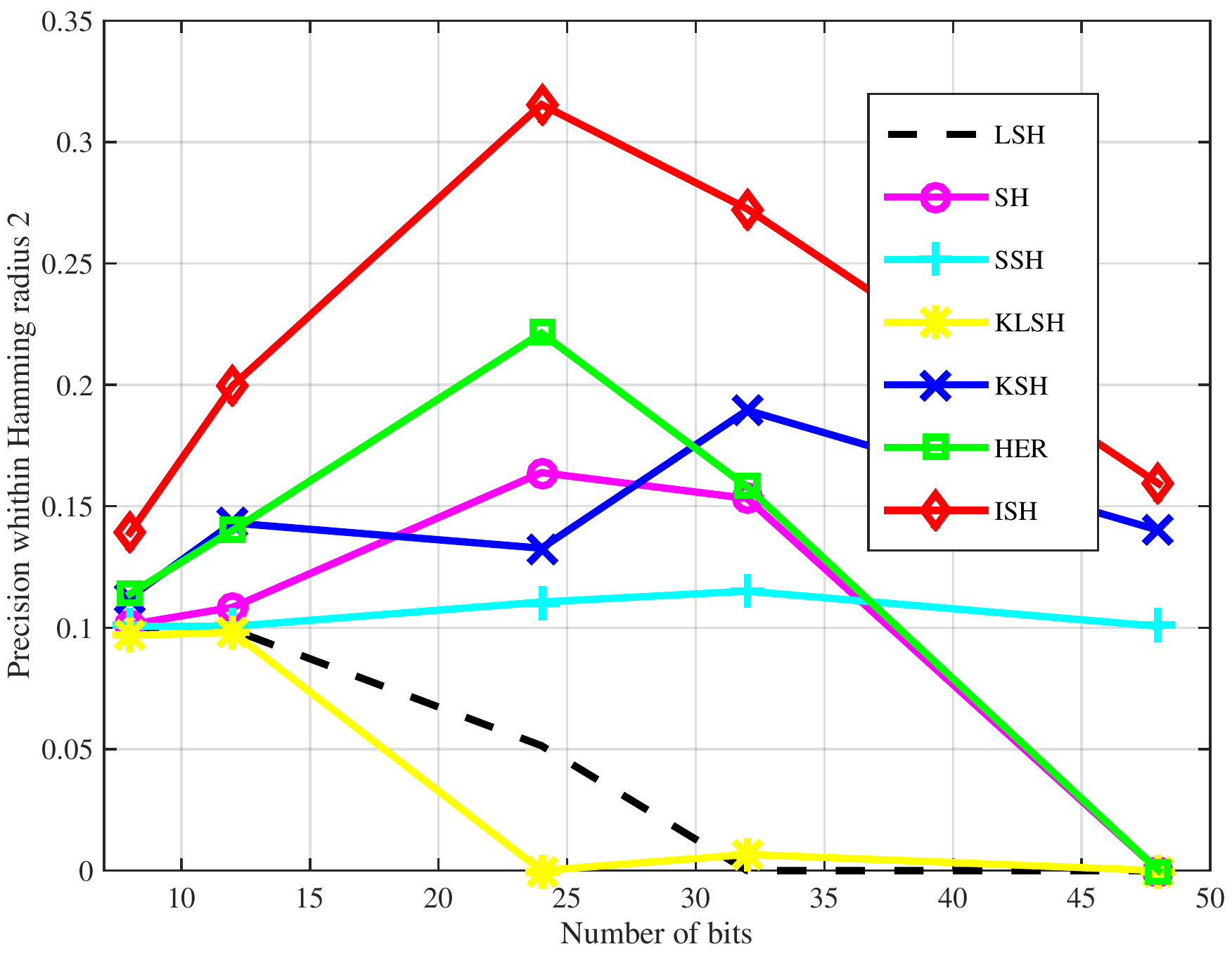, height=1.5in, width=1.8in }
    }
    \vspace{-7mm}
\caption{Mean precision curves of Hamming radius 2 in $24$ bits on CIFAR-10 dataset. The curves of LSH, SH, SSH are randomly sampled. The figure is best viewed in color.}
\label{fig:rd_cifar}
  \end{minipage}
\end{figure*}



%% file: sec5.tex
\section{Experiments}
\vspace{-2mm}
In this section, we evaluate the effectiveness of the proposed Image Set Hashing (ISH) method on two well-known benchmarks, CIFAR-10~\footnote{http://www.cs.toronto.edu/~kriz/cifar.html} and TV-series, i.e., Big Bang Theory~\footnote{https://cvhci.anthropomatik.kit.edu/~baeuml/datasets.html}. We also conduct extensive comparison studies with state-of-the-art methods, including Locality Sensitive Hashing (LSH)~\cite{Indyk_98}, Spectral Hashing (SH)~\cite{Weiss_08}, Kernelized LSH (KLSH)~\cite{Kulis_nips2009}; Semi-Supervised Hashing (SSH)~\cite{Wang_icml10} and supervised methods, Kernel-Based Supervised Hashing (KSH)~\cite{Liu_cvpr12} and Hashing across Euclidean space and Riemannian
manifold (HER)~\cite{Li_cvpr15}. For the competing techniques, we adopted the publicly released codes of SH, KLSH, KSH and HER in our experiments. All the experiments set the number of iterations $T$ to $15$ and are performed on the MATLAB platform using a Six-Core Intel Xeon Processor with 2.8 GHz CPU and 48GB RAM. 

\subsection{Experiment on CIFAR-10}
\vspace{-1mm}
We compare the performance for different hashing techniques on the CIFAR-10 dataset. As a labeled subset of the $80$M tiny images, the CIFAR-10 dataset consists of a total of $60$K color images, each of which has the size of $32 \times 32$ resolution. The dataset contains $6$K image samples with ten object categories. To evaluate the performance, we uniformly and randomly sample images from each category to form a total of $195$ image sets, each of which contains about $25 \sim 50$ images for the training process (q and r two parts), $100$ image sets as query in testing and $1577$ image sets for the testing database in KLSH, KSH, HER, ISH methods. To test the LSH, SH and SSH methods, we randomly select $1$K images as queries and the remaining as database samples. For feature representation, each image is represented as a 512-dimensional GIST feature vector~\cite{Oliva_ijcv01}.




We evaluate two test scenarios: Hamming ranking and hash look-up. For the retrieval strategy based on Hamming ranking, the performance measured by mean average precision (MAP) is reported in Table~\ref{tb:cifar} for the hash codes with the length from $8$-bit to $48$-bit. In addition, Fig.~\ref{fig:re_cifar} and Fig.~\ref{fig:pr_cifar} show the mean recall and precision curves from different number of returned search samples when using 24-bit hash codes. Clearly, the proposed ISH method produces higher quality of Hamming embedding since it significantly outperforms the competing methods in terms of precisions, recalls, and MAPs. In general, the methods using set information often provide better performance than those based on P2P settings. For instance, the HER method generates the second best MAPs for most of the test cases. The relative performance gains in MAP ranges from $6\%$ to $23.6\%$ compared to the HER method. Such performance gains confirm the value of exploring statistical and clique-based structural information for hash function design. In Fig.~\ref{fig:rd_cifar}, we also evaluate the results using the hash look-up table strategy by showing the precision curve within Hamming radius 2 for hash codes from $8$-bit to $48$-bit. Again the proposed ISH method achieved the best precisions across all the cases.

\subsection{Experiment on TV-series}
The Big Bang Theory (BBT) video (image set) benchmark was collected by~\cite{Bauml_cvpr13} and contains in $3341$ face videos from $1\sim6$ episodes of season one. The dataset includes around $5\sim8$ main cast characters and has multiple characters at the same full-view scene shot. Even though most of the scenes are taken in indoors, it is still extremely challenging since the resolution of faces regions are quite small with an average size of 75 pixels. In the experiments, we use the provided face features, which are extracted from face videos by block Discrete Cosine Transformation (DCT) feature. In this way, each face is represented by a $240$-dimensional DCT feature vector.


In the experiments, we have two different settings. For the first setting, we follow the setting used in~\cite{Li_cvpr15} and apply still images for the q part in training and query in testing process and denote the setting as $\textrm {ISH}^0$. The second setting indicated as ISH, we have $150$ image sets for q and r two parts in the training process, respectively. For query in testing, we use $100$ image sets and the remaining image sets for database. The setting can completely utilize statistical and structural information. For the comparison, the first group of compared methods consists of seven point-to-point (P2P) hash methods, i.e, LSH, ITQ, SH, Discriminative Binary Codes (DBC)~\cite{Rastegari_eccv12} , SSH, MM-NN~\cite{Masci_pami13}, and KSH (point). In addition, we generate kernels for image sets (represented by covariance matrices) and employ them as input for the KLSH, KSH (set), and HER methods. Thus, we have the second group of methods that uses kernels for image sets. Such a modification can be used to further justify the advantage of explore the structural and statistical information of image sets. The performance evaluated by MAPs for the compared methods is shown in Table~\ref{tb:tv}. We vary the number of hash bits from $8$ to $128$ bits. As we can see, the method ($\textrm {ISH}^0$) generates the second best MAPs when structural information is ignored in the q training and the query testing process. Moreover, ISH method achieved the best performance for all the tested cases when both of statistical and structural information are considered. This evidence shows that the structural information can help to generate more robust and discriminative hash codes. 

\begin{table*}[t]
\vspace{-1mm}
\scriptsize
\caption{The evaluation results measured by Mean Average Precision on the on video (image set) benchmark TV-series (BBT). The length of hash codes ranges from $8$-bit to $128$-bit. Besides the proposed ISH, the first group of compared methods consists of seven P2P hashing methods, i.e., LSH, ITQ, SH, DBC, SSH, MM-NN, and KSH. The second group of compared methods include three modified techniques, i.e., KLSH, KSH, and HER, that use image set information as input.} 
\label{tb:tv}
\begin{center}
\setlength{\tabcolsep}{16pt}
\begin{tabular}{|p{2.2cm}||p{1cm}|p{1cm}|p{1cm}|p{1cm}||p{1.1cm}|}
\hline
         & &\multicolumn{4}{c|} {TV drama: the Big Bang Theory}  \\ \hline\hline
          Method  &  8 bits      &    16 bits &  32 bits      &  64 bits       &  128 bits           \\ \hline \hline
         LSH~\cite{Indyk_98}     &  $0.2109$          & $0.2086$            & $0.2092$       &  $0.1963$  &	$0.1994$    \\
         ITQ~\cite{Gong_cvpr11}    &  $0.2935$          & $0.3025$            & $0.2989$       &  $0.3029$  &	$0.3060$    \\
         SH~\cite{Weiss_08}    &  $0.2377$          & $0.2652$            & $0.2665$       &  $0.2623$  &	$0.2673$    \\
         DBC~\cite{Rastegari_eccv12}    &  $0.4489$          & $0.4495$            & $0.4235$       &  $0.4005$  &	$0.3867$    \\
         SSH~\cite{Wang_icml10}    &  $0.2716$          & $0.2855$            & $0.2662$       &  $0.2584$  &	$0.3003$    \\
         MM-NN~\cite{Masci_pami13}    &  $0.3752$          & $0.3955$            & $0.4664$       &  $0.5124$  &	$0.4922$   \\
         \hline\hline
         KLSH~\cite{Kulis_nips2009}    &  $0.2450$          & $0.2498$            & $0.2381$       &  $0.2256$   &	$0.2325$    \\
         KSH (point)    &  $0.4090$          & $0.4366$            & $0.4454$       &  $0.4567$    &	$0.4604$   \\
         KSH~\cite{Liu_cvpr12} (set)    &  $0.4590$          & $0.4619$            & $0.4534$       &  $0.4685$    &	$0.4631$   \\
         HER~\cite{Li_cvpr15}    &  $0.4606$          & $0.5049$            & $0.5227$       &  $0.5490$    &	$0.5539$    \\
         $\textrm {ISH}^0$    &  $0.4833$          & $0.5279$          & $0.5359$    &  $0.5501$    &	$0.5712$ \\
         \hline
         ISH    &  $\textbf{0.5018}$          & $\textbf{0.5592}$          & $\textbf{0.5864}$    &  $\textbf{0.6007}$    &	$\textbf{0.6280}$ \\                           
\hline
\end{tabular}
\end{center}
\end{table*}

\begin{figure*}
    \hspace{-1mm}
  \begin{minipage}{.3\textwidth}
    {
    \hspace{-2.5mm}
    \epsfig{file=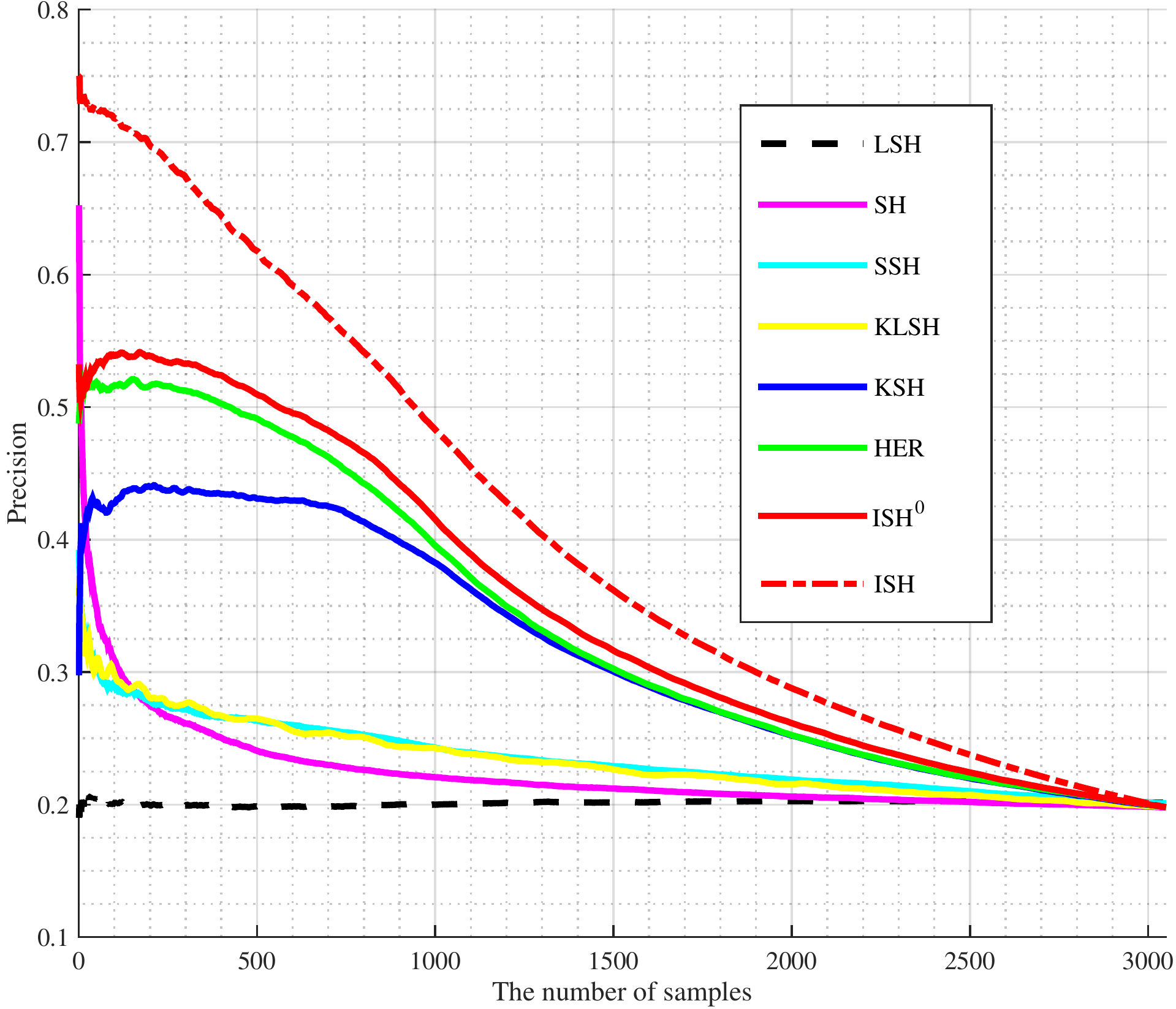, height=1.7in, width=1.8in }
    }
    \vspace{-5mm}
  \caption{The evaluation results by mean precision curves for Hamming ranking using $128$ bits on the BBT video dataset. The figure is best viewed in color.}
  \label{fig:tv_pr}
  \end{minipage}
    \hspace{2.5mm}
  \begin{minipage}{.3\textwidth}
		{
		\vspace{-3mm}
		\epsfig{file=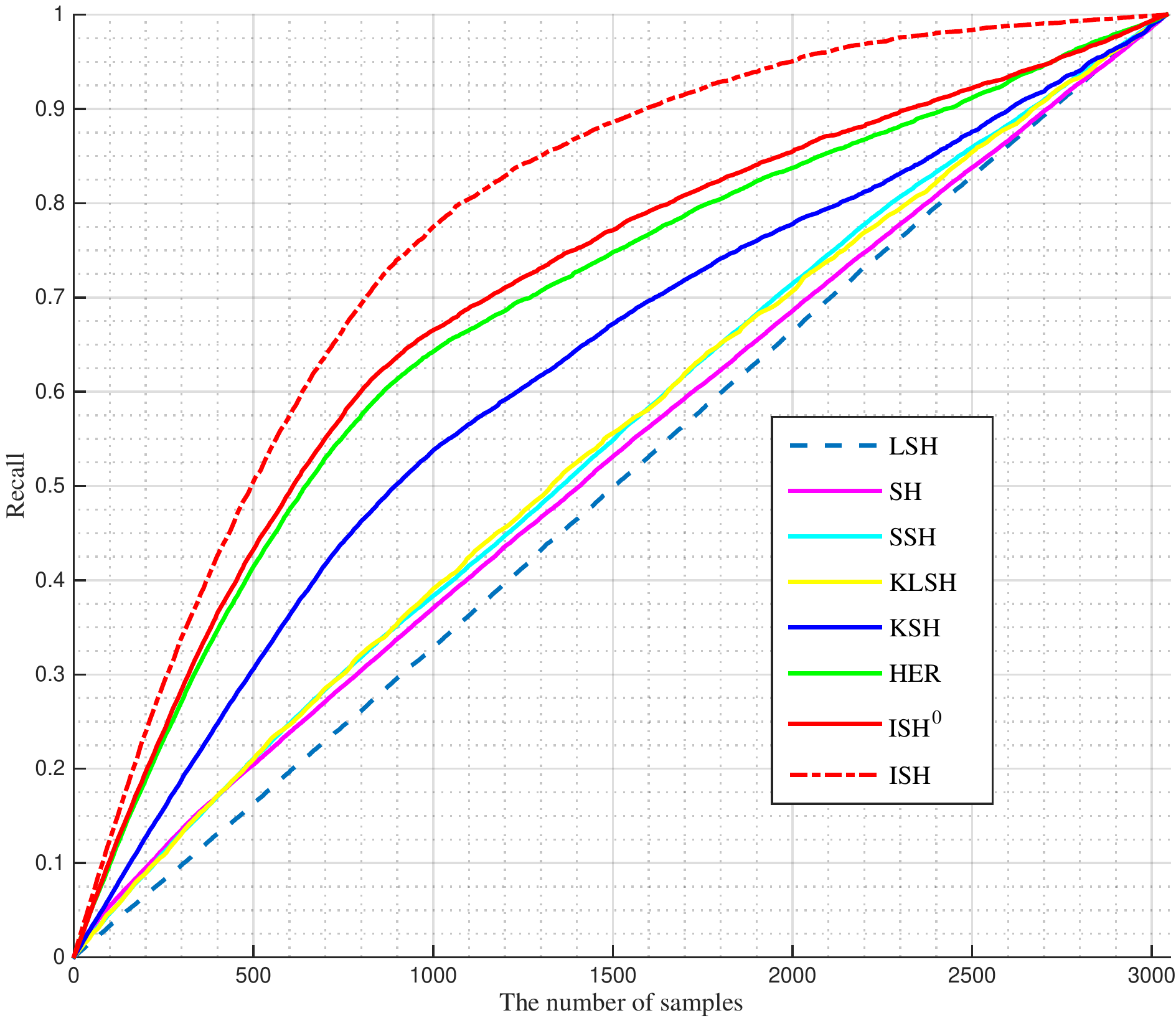, height=1.71in, width=1.8in }

		}
  \caption{The evaluation results by mean recall curves for Hamming ranking using $128$ bits on the BBT dataset. The figure is best viewed in color.}
  \label{fig:tv_rc}  
  \end{minipage}
  \hspace{3mm}
  \begin{minipage}{.3\textwidth}
    {
     \epsfig{file=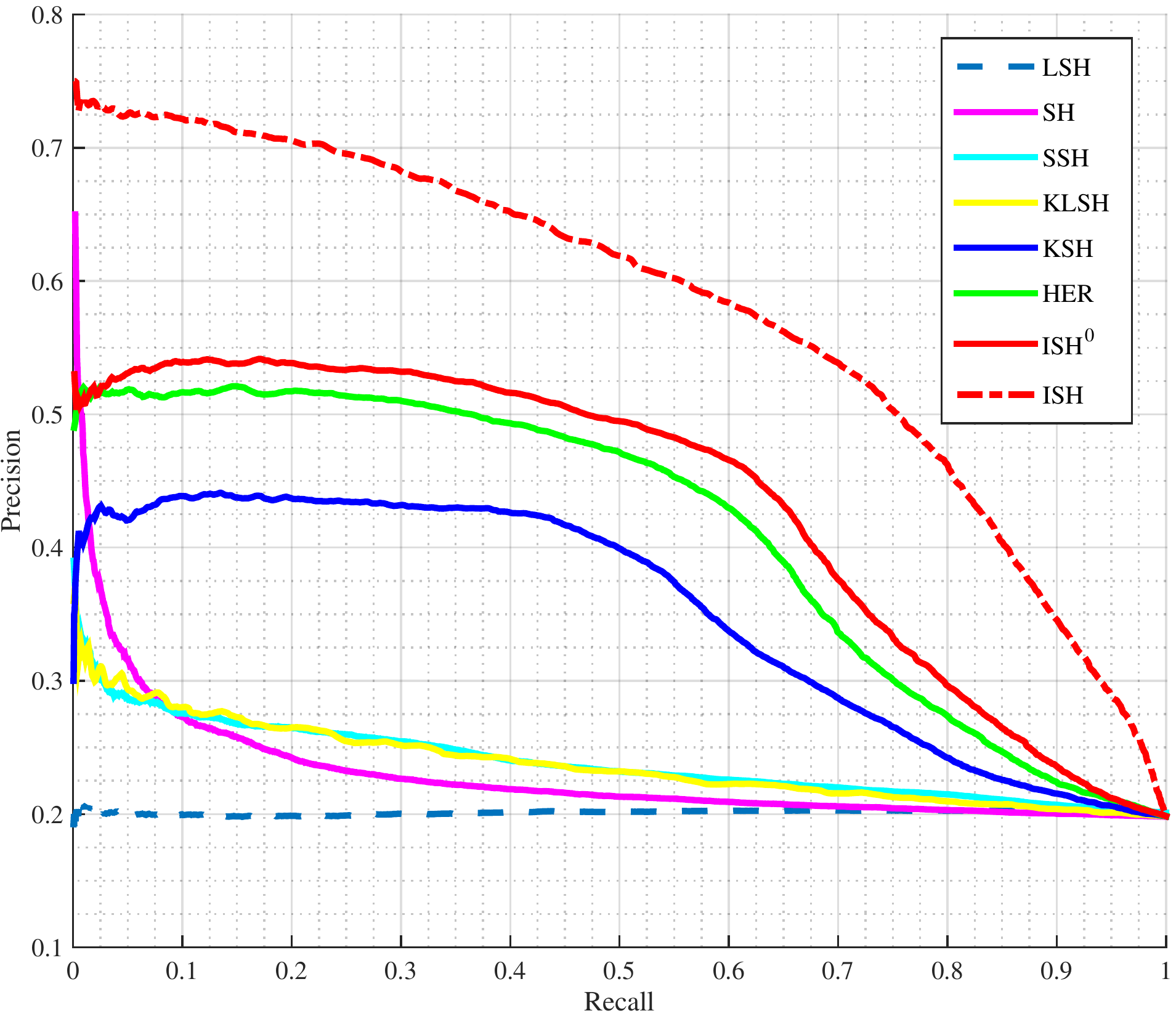, height=1.7in, width=1.8in }}
  \vspace{-4mm}
  \caption{The evaluation results by mean precision-recall curves for Hamming ranking using $128$ bits on the BBT dataset. The figure is best viewed in color.}
  \label{fig:tv_pr_rc}
  \end{minipage}
\end{figure*}

In addition, we use $128$-bit hash codes and show the performance curves for the seven compared methods. Fig.~\ref{fig:tv_pr}, Fig.~\ref{fig:tv_rc} and Fig.~\ref{fig:tv_pr_rc} show the precision, recall, and precision-recall curves, respectively. From these results, the proposed ISH achieves the best performance compared to all the baseline techniques, including both P2P methods and modified S2S methods. The underlying reason lies in that the ISH method can simultaneously capture the statistical (covariance matrix) and structural (graph kernel) information, i.e., combination of weak learners, to generate each hash code. Hence, it captures the most intrinsic characteristics within complicated variations of face images for the same subject. Moreover, if we compare the general performance between the P2P and S2S settings for the KSH method, we can observe that the S2S hashing continuously attains higher search accuracy. It further confirms the advantage of using set information. However, simply employing the covariance matrix as inputs limits the performance improvements. In summary, with the S2S setting and fully explore statistical and structural information, the proposed ISH yields significant performance gain across all the experiments.

%% file: sec6.tex
\section{Conclusion}
\vspace{-1mm}
In this paper, we have presented a set-to-set (S2S) ANN search problem and proposed to learn the optimal hash codes for image sets by simultaneously exploiting the statistical and structural information. The key idea is to transform the image sets into a high dimension space where each of image set can be characterized by a graph kernel and statistical measurement. Specifically, the proposed Image Set Hashing (ISH) method captures the relations among each image set based on a clique structure. Meanwhile, it considers the content information using covariance matrix representation. As a result, the proposed S2S hashing achieves a robust and discriminative representation for searching datapoint sets. We have conducted extensive experiments on two well-known benchmarks. The experimental results have demonstrated the effectiveness of the proposed ISH method by showing superior performance over several representative competing approaches. For our future work, we will further extend the proposed method for more general set matching and search problems. In addition, we will consider incorporating sparse representation for set data to further boost the performance.